\RequirePackage[hyphens]{url} % break urls

\documentclass[runningheads]{llncs}

\PassOptionsToPackage{hyphens}{url}
\usepackage{hyperref} % break urls

\usepackage{cite}
\usepackage{lscape}
\usepackage{amsfonts} 
\usepackage{graphicx}
\usepackage{amsmath}
\usepackage{marvosym} % for email symbol \Letter
\usepackage{soul, color}
\begin{document}
% Abstract registration: 22 February 2024
% Submission deadline:   7 March 2024
% early acceptance notification: 17 May 2024
% rebutal: 20 May 2024
% Notification of paper: 17 June
% Camera-ready paper submission: 8 July
% Conference: 6-8 October
\title{This actually looks like that: Proto-BagNets for local and global interpretability-by-design}
\titlerunning{Proto-BagNets for local and global interpretability-by-design}

% \author{Anonymous Authors}

\author{Kerol Djoumessi\inst{1,2}\textsuperscript{(\Letter)}\orcidID{0009-0004-1548-9758}
\and 
Bubacarr Bah\inst{3} 
\and 
Laura Kühlewein\inst{4} 
\and 
Philipp Berens\inst{1,2}\textsuperscript{(\Letter)}\orcidID{0000-0002-0199-4727} \and
Lisa Koch\inst{1,2,5}\textsuperscript{(\Letter)}\orcidID{0000-0003-4377-7074}
}
% 1 {Kerol, Djoumessi}
% 2 {Bubacarr, Bah}
% 3 {Laura, Kühlewein}
% 4 {Philipp, Berens}
% 5 {Lisa, Koch}

% \authorrunning{Anonymous et al.}
\authorrunning{K. Djoumessi et al.}
% First names are abbreviated in the running head.
% If there are more than two authors, 'et al.' is used.

% \institute{Anonymous \\ ** \\ ** \\ ** \\ ** \\ **  }

\institute{Hertie Institute for AI in Brain Health, University of T{\"u}bingen, Germany \\ \email{\{kerol.djoumessi-donteu, philipp.berens\}@uni-tuebingen.de} 
\and 
T{\"u}bingen AI Center, University of T{\"u}bingen, Germany 
\and 
Medical Research Council Unit The Gambia at London School of Hygiene and Tropical Medicine 
\and 
University Eye Clinic, University of Tübingen, Germany 
\and 
Department of Diabetes, Endocrinology, Nutritional Medicine and Metabolism UDEM, Inselspital, Bern University Hospital, University of Bern, Switzerland \\
\email{\{lisa.koch\}@unibe.ch} 
}

\maketitle              % typeset the header of the contribution
%\the\textwidth 
\begin{abstract}
    Interpretability is a key requirement for the use of machine learning models in high-stakes applications, including medical diagnosis. 
    Explaining black-box models mostly relies on post-hoc methods that do not faithfully reflect the model's behavior. 
    As a remedy, prototype-based networks have been proposed, but their interpretability is limited as they have been shown to provide coarse, unreliable, and imprecise explanations. 
    In this work, we introduce Proto-BagNets\footnote{Code available at \url{https://github.com/kdjoumessi/Proto-BagNets}}, an interpretable-by-design prototype-based model that combines the advantages of bag-of-local feature models and prototype learning to provide meaningful, coherent, and relevant prototypical parts needed for accurate and interpretable image classification tasks. 
    We evaluated the Proto-BagNet for drusen detection on publicly available retinal OCT data. The Proto-BagNet performed comparably to the state-of-the-art interpretable and non-interpretable models while providing faithful, accurate, and clinically meaningful local and global explanations.    

\end{abstract}
\section{Introduction}
    For adopting deep learning models in safety-critical applications such as medical diagnosis, it is crucial that users can understand why a model produced a specific output  \cite{rudin2019stop}. This form of interpretability is usually obtained either through post-hoc explanations of black-box models \cite{zhou2016learning} or through architectural design \cite{chen2019looks, donteu2023sparse}.  
    Post-hoc methods \cite{zhou2016learning, selvaraju2017visual} interpret an approximation of the true decision mechanism \cite{jacovi2020towards} through saliency maps. These highlight the most discriminating regions in the input, but often provide inaccurate and unfaithful explanations \cite{adebayo2018sanity, rudin2019stop}. 
    To remedy this, several approaches have been proposed with structurally built-in interpretability, such as bag-of-local-features models (BagNets) \cite{brendel2019approximating}, concept-based models \cite{koh2020concept}, and prototype-based models \cite{chen2019looks}. 

    The BagNet \cite{brendel2019approximating} is an implicitly patch-based interpretable-by-design model with a small receptive field, where predictions solely rely on local evidence.
    Its recent modification \cite{donteu2023sparse} provides sparse and fine-grained local class activation maps, but does not allow humans to gain a global understanding of the model's decision.
    Concept-based models \cite{koh2020concept} follow a case-based reasoning process where high-level representations of the data (concepts) are learned and used to classify new images. 
    Prototype-based networks \cite{chen2019looks} can be seen as a special case of concept-based models, in which learned concepts are replaced by the representative training image parts (prototypes) to improve interpretability. 
    %Similarities to learned prototypes are used to classify new examples in prototype-based models. 
    In these models, similarities to the learned prototypes are used to classify new examples.
    Explanations can be obtained during inference by highlighting, for a query image, its prototypical parts most similar to each learned prototype, thus providing both local explanations thanks to the similarity map and global explanations through the visualization of the learned prototypes.
    ProtoPNet \cite{chen2019looks}, the first prototype-based network, has gained considerable attention due to its easy-to-understand architecture and high-level reasoning process close to that of humans in solving complex tasks. 
    Although numerous variants have been proposed to improve its performance and interpretability \cite{barnett2021case, kim2021xprotonet, huang2023evaluation, gautam2023looks}, applications in medical imaging remain relatively limited.
    This may in part be due to the fact that the interpretability of prototype-based models is more limited than appears at first glance, as it has been shown that they do not actually provide faithful explanations \cite{xu2023sanity, gautam2023looks}.

    We propose Proto-BagNet, an interpretable-by-design prototype-based model that combines the local and fine-grained interpretability of BagNet with the global interpretability of prototype learning. We integrated recent advances in training prototype-based models and proposed an additional prototype diversity constraint. We evaluated our model for detecting drusen lesions on Optical Coherence Tomography (OCT) images and showed that the Proto-BagNet preserves high predictive performance while providing faithful, clinically meaningful, and precise explanations. Our model explanations accurately localized drusen both in the learned prototypes and query test images. 
        
\section{Developing a faithful prototype-based network}
    \begin{figure}[t]
        \centering
        \includegraphics[width=.9\textwidth]{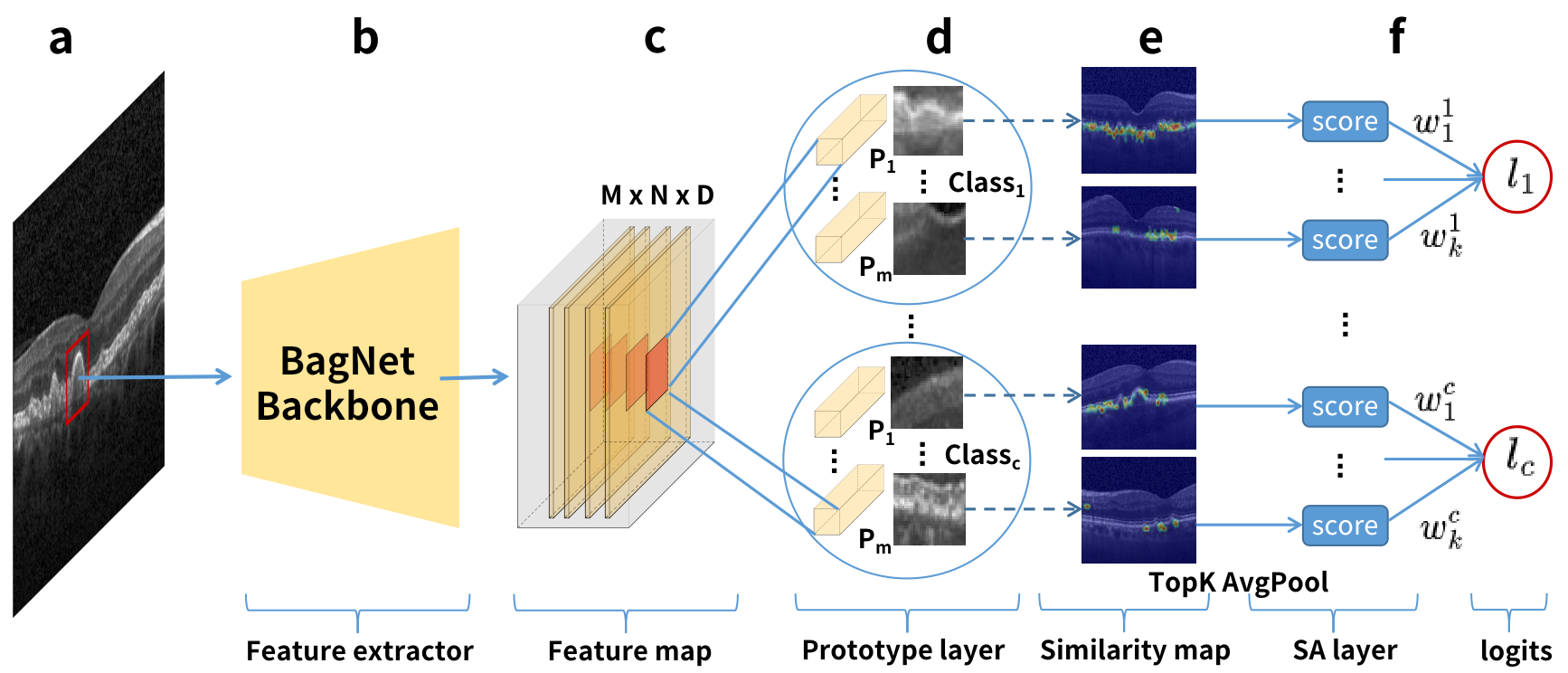}
        %\vspace{-0.9cm}
        \caption{Architecture of the Proto-BagNet. \textbf{(a)} Example OCT B-scan image. The red patch illustrates the small receptive field of \textbf{(b)} the BagNet backbone. \textbf{(c)} Feature map and \textbf{(d)} Prototype layer with $m$ prototypes per class. 
        \textbf{(e)} Resulting similarity maps from each prototype to the input. \textbf{(f)} The soft aggregation layer aggregates the average top-k scores from each similarity map into their allocated categories for classification.}
        \label{fig1}
        %\vspace{-0.6cm}
    \end{figure}
    
    \subsection{Baseline ProtoPNet model}
        \label{baseline}
        We built on the ProtoPNet \cite{chen2019looks} as a baseline, which consists of three main components: a backbone feature extractor $f$, a prototype layer $g_p$, and a classification layer $h$. 
        Given an input image $\textbf{X} \in \mathbb{R}^{H \times W \times C}$ (with height $H$, width $W$, and number of channels $C$), the backbone first extracts a meaningful feature representation $\mathbf{Z}=f(\mathbf{X}) \in \mathbb{R}^{M \times N \times D}$. 
        The prototype layer $g_\mathbf{p}$ consists of $b=m \times c$ learnable prototypes $\mathbf{P} = \{\mathbf{p}^i_j \in \mathbb{R}^{H_P \times W_P \times D}\}_{j=1}^m$ (typically $H_P = W_P=1$) where $m$ denotes the number of prototypes per class, $c$ the number of classes, and $\mathbf{p}^i_j$ the $j$-th prototype of class $i$. 
        For each prototype $\mathbf{p}^i_j$, the prototype layer computes a similarity map $ M_{\mathbf{p}_j^i}^{\mathbf{x}} = \mbox{Sim}(\mathbf{Z}, {\mathbf{p}_j^i}) \in \mathbb{R}^{M \times N}$. The similarity score between a prototype $\mathbf{p}^i_j$ and the feature vector $\mathbf{z}^{(h,w)} \in \mathbf{Z}$ is defined as $s_{i,j}^{(h,w)}= \log \big((d_{i,j}^{h,w} + 1) / (d_{i,j}^{h,w} + \epsilon) \big)$, where $d_{i,j}^{h,w} = \Vert \mathbf{p}_j^i -  \mathbf{z}^{(h,w)}\Vert^2_2$. % is the squared Euclidean distance between $\mathbf{p}^i_j$ and $\mathbf{z}^{(h,w)}$.            
        The similarity maps contain positive scores indicating where and to what extent prototypes are present in an image.
        ProtoPNet uses the highest value of the similarity map $g_{\mathbf{p}_j^i} = \max (M_{\mathbf{p}_j^i}^{\mathbf{x}})$ as the final similarity score between $\mathbf{p}_j^i$ and $\mathbf{X}$, indicating how strong the prototype $\mathbf{p}_j^i$ is present in $\mathbf{X}$.
        Finally, the $b$ similarity scores from the prototype layer $g_p$ are aggregated in the fully connected layer $h$ to generate the final classification logits. 
        To make the prototypes visualizable as specific prototypical parts of a sample, the learned prototypes are replaced with the closest feature representation from real training images to ensure interpretability.        
        
        % How do the explanations work
        ProtoPNet explains its predictions for a given image (``local explanation'') by (1) visualizing the similarity map for each prototype $\mathbf{p}_j^i$ and (2) by computing the smallest bounding box enclosing the $95th$ percentile of all similarity values \cite{chen2019looks}, providing the corresponding cropped region as the most similar part to the learned prototype to imply \textit{'this part of the input looks like that learned prototype'}. 
        The same approach is used to provide explanations of the concepts learned by the model (``global explanations'') by cropping the prototypes from the most similar training image.
        %%% issues with the explanation      
        However, ProtoPNet provides only coarse-grained similarity maps due to the large receptive field size of the model \cite{gautam2023looks}. Furthermore, the explanation is not faithful to the model, as the cropped area does not correspond to the model's receptive field. 
        As a result, ProtoPNet provides both imprecise local and global explanations of its decisions. These issues are likely to be shared by all prototype-based models derived from ProtoPNet \cite{xu2023sanity}. 

    % \vspace{-1.5pt}
    \subsection{Enhancing interpretability with the BagNet backbone}   
        \label{backbone}
        As prototypes are learned in the feature space, the backbone feature extractor plays a crucial role for interpretability. 
        It implicitly determines the size of the learned prototypes through its receptive field and thus the size of the explanation. 
        The ProtoPNet and its variants use classical architectures such as ResNet-50 \cite{chen2019looks, kim2021xprotonet, huang2023evaluation}, resulting in large receptive fields (e.g. $427 \times 427$ for a ResNet-50 backbone) with variable explanation size.
        Here, we propose to replace the feature extractor with a BagNet architecture \cite{brendel2019approximating, donteu2023sparse} (Fig. \ref{fig1}b), leading to a model we called Proto-BagNet (Fig. \ref{fig1}). 
        The feature map $\mathbf{Z}=f(\mathbf{X})$ is extracted from the BagNet's penultimate layer (Fig. \ref{fig1}c, typically $D=2048$). 
        This architecture leads to a small fixed receptive field and prototypes of size $r \times r$ independent of the input size. 
        It also allows for higher-resolution feature and prototype similarity maps, and can therefore provide localized, fine-grained explanations.

    % \vspace{-1.5pt}    
    \subsection{Integrating recent advances in training prototype-based models}
    % \vspace{-1.5pt}
        In addition, we implemented recent advances in prototype-based networks training \cite{huang2023evaluation, barnett2021case, kim2021xprotonet}. 
        (1) To prevent prototypes of one class from contributing to the prediction of other classes, we replaced the fully connected classification layer of ProtoPNet with a soft aggregation (SA) layer (Fig. \ref{fig1}f) \cite{huang2023evaluation}, which aggregates the prototypes' similarity scores only in their assigned classes, setting weights between classes to zero. 
        (2) To enable the model to consider multiple image regions for the classification task instead of considering only the region with the highest score of each similarity map as in ProtoPNet, we considered the top-$k$ scores through average pooling as $g_{\mathbf{p}j^i} = \mbox{AvgPool} \big(\mbox{topk}(M_{\mathbf{p}_j^i}^\_)\big)$ \cite{barnett2021case}. 
        Thus, our similarity map indicates to what extent a prototype is present on average in the $k$ most similar prototypical parts of the input. 
        (3) We regularized the prototype layer by adding a sparsity constraint to each similarity map as in \cite{kim2021xprotonet, donteu2023sparse}, to constrain activation to discriminative input regions.  
        (4) Finally, as we noticed redundant prototypes (often extracted from the same training image, see Suppl. Fig. 1), we introduced a dissimilarity loss (see below) to prevent the network from learning duplicate prototypes while promoting their coherence and uniqueness. 
        Thus, the total loss function was:
        \begin{equation*}
            \mathcal{L} = \mathcal{L}_{{ce}} + \underbrace{\lambda_{{clst}} \mathcal{L}_{{clst}} + \lambda_{{sep}} \mathcal{L}_{{sep}}}_\text{ProtoPNet} + \lambda_{L1,c} \mathcal{L}_{L1,c} + \lambda_{L1,s} \mathcal{L}_{L1,s} - \lambda_{diss} \sum_{{p}_i, p_j} \left \Vert \mathbf{p}_i - \mathbf{p}_j \right \Vert^2 
            \label{eq:loss}
        \end{equation*}        
        Here, $\mathcal{L}_{\text{ce}}$ is the cross-entropy loss; $\mathcal{L}_{clst}$ and $\mathcal{L}_{sep}$ the cluster and separation losses from ProtoPNet \cite{chen2019looks}, $\mathcal{L}_{L1,c}$ is the $\ell_1$ regularization of the classification layer as in \cite{chen2019looks}; $\mathcal{L}_{L1,s}$ regularizes the similarity maps \cite{kim2021xprotonet, donteu2023sparse}. 
        Finally, $\sum_{p_i, p_j} \left \Vert \mathbf{p}_i - \mathbf{p}_j \right \Vert^2$ is our proposed dissimilarity loss with $\mathbf{p}_i, \mathbf{p}_j \in \mathbf{P}, i \ne j$.  %(see App.\ref{training_proto})
        
        Our architectural changes alongside with the modified loss function led to an interpretable-by-design prototype-based model (Proto-BagNet) that is easier to interpret, relying on the small receptive field of the model for predictions and explanations. 
        In addition, Proto-BagNet provides accurate local and global explanations (see Sec. \ref{localize_explanations}, and \ref{topk_implication}) in the form of \textit{'this part of the input actually looks like that learned prototype'}. 
        
\section{Results}
    \subsection{Dataset}
        We used a publicly available, anonymized dataset \cite{kermany2018identifying} consisting of retinal OCT B-scans from patients with various diseases (drusen, DME, CNV). We focused on the binary task of drusen detection and filtered out images with DME and CNV diagnoses, as well as low-resolution images (width $<496$). To counter the class imbalance, we removed half of the healthy images, leading to a dataset of $34,962$ images ($8,616$ drusen, $26,346$ healthy).
        We split the resulting dataset into training ($80\%$) and validation ($20\%$) sets, preserving the imbalance proportion ($73\%$ vs $27\%$) and ensuring that all B-scans from each patient were assigned to the same set. 
        We then used the separate test set included in the dataset for evaluation, consisting of $250$ healthy and $248$ drusen images ($51\%$ vs $49\%$), reflecting the high variability of drusen prevalence according to age group \cite{silvestri2005prevalence}. 
        All images were resized to $496 \times 496$ and normalized by the mean and standard deviation of the training set. 
        To evaluate the relevance of the learned prototypes, an experienced in-house ophthalmologist provided detailed drusen annotations on a selection of $40$ test images.  
        
    \subsection{Proto-BagNet yields good accuracy on drusen detection} % classification performance  
        We first evaluated the classification performance of our method for a clinically relevant binary task of detecting patient's OCT-B scans with drusen lesions (lipid deposits under the retina \cite{fleckenstein2021age}), characteristic of age-related macular degeneration and diabetic retinopathy \cite{kermany2018identifying, fleckenstein2021age}.
        For Proto-BagNet, we configured the backbone feature extractor (BagNet model) to a receptive field size $r=33$ as in \cite{donteu2023sparse}.
        Hyperparameters including regularization coefficients, data augmentation, and the number of prototypes were optimized on the validation dataset using a grid search, while $\lambda_{sep}$ and $\lambda_{clst}$ were set as in \cite{chen2019looks}. 
        Based on the validation performance, we set $k=5$ considering the average top-$5$ and used $m=5$ prototypes per class, which lead to a total of $b=10$ prototypes. 

        \begin{table}[b]
            \centering
            \caption{Classification performance for drusen detection on validation and test sets.}
            \begin{tabular}{l | c c c c | c c c c}
                \hline
                & \multicolumn{4}{c}{Validation set} & \multicolumn{4}{c}{Test set} \\
                 & Accuracy & AUC & Recall & Precision & Accuracy & AUC & Recall & Precision \\
                \hline
                ResNet-50 & $0.991$ & $0.999$ & $0.982$ & $0.986$ & $0.994$ & $0.999$ & $0.992$ & $0.996$ \\
                dense BagNet & $0.990$ & $0.999$ & $0.978$ & $0.985$ & $0.988$ & $0.999$ & $0.976$ & $0.999$ \\
                \hline
                ProtoPNet & $0.987$ & $0.996$ & $0.975$ & $0.974$ & $0.998$ & $0.999$ & $0.996$ & $0.999$ \\
                Proto-BagNet & $0.978$ & $0.990$ & $0.935$ & $0.981$ & $0.968$ & $0.992$ & $0.940$ & $0.996$ \\
                \hline
            \end{tabular}
            \label{tab:classification_results}
        \end{table} 

        We compared Proto-BagNet against ProtoPNet with a ResNet-50 backbone \cite{chen2019looks} and with non-prototype classification networks such as a dense BagNet \cite{donteu2023sparse} and ResNet-50 \cite{He_2016_CVPR}.         
        We followed the same training procedure for ProtoPNet (with $\lambda_{L1,s} \, $=$\, \lambda_{diss}$=$\, 0$, $K$=$1$) and Proto-BagNet (with $\lambda_{L1,s}$=$\, 4 \cdot 10^{-2}, \, \lambda_{diss}$=$\, 5 \cdot 10^{-3}$, $K$=$\, 5$), as well as for dense BagNet and ResNet-50. 
        Our Proto-BagNet performed comparably to the state-of-the-art models (Tab.\,\ref{tab:classification_results}, see confidence intervals in Suppl. Tab. 1), showing that our modifications towards better interpretability did not substantially impair classification performance.   
        
    \subsection{Proto-BagNet provides understandable localized explanations}     
        \label{localize_explanations}
        \begin{figure}[t]
            \centering
            \includegraphics[width=\textwidth]{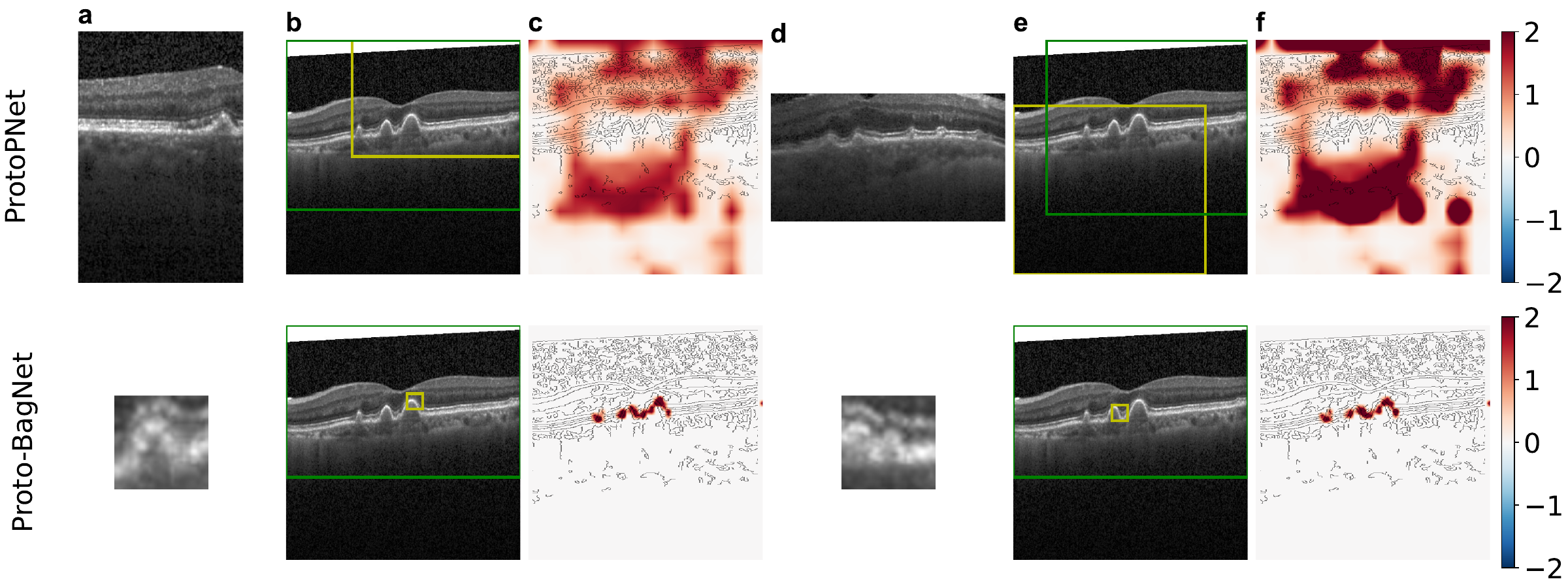}
            %\vspace{-0.9cm}
            \caption{Example explanations of ProtoPNet and Proto-BagNet. \textbf{(a,d)} show two learned prototypes with the highest classification weights. Proto-BagNet's prototypes were magnified for visualization only. \textbf{(b,e)} show bounding boxes around regions of highest activation using the visualization technique provided by ProtoPNet (green) and the model's receptive field (yellow). \textbf{(c,f)} show prototypes activations of the query image.}
            \label{fig2}
            %\vspace{-0.6cm}
        \end{figure}
        
        To qualitatively assess explanations provided by our model, we visualized the two learned prototypes with the highest classification weights for our Proto-BagNet as well as ProtoPNet (Fig.\,\ref{fig2}a,d). 
        ProtoPNet learned very large prototypes covering almost the entire retina, while Proto-BagNet learned prototypes of small regions of interest with fixed sizes corresponding to its receptive field. 
        For a query image, we then displayed bounding boxes (Fig. \ref{fig2}b,e) around the most similar prototypical part to the learned prototypes.
        For both models, we first computed the explanation of the prototypical part as the receptive field around the location of the highest prototype similarity (yellow boxes). 
        Due to the large receptive field in ProtoPNet, the explanations were not informative, while the Proto-BagNet yielded small localized patches of the same size as the prototypes themselves.
        We then computed the explanation as the bounding box around the $95$th percentile of the similarity map as in \cite{chen2019looks}, which is not faithful to the model's predictions, as it usually leads to large and similar explanations (green bounding boxes, Fig. \ref{fig2}b,e). 
        In both models, this again led to large bounding boxes around the entire retina, indicating that such prototype explanations may not be useful when evidence may be spread in an image (see similarity maps in Fig. \ref{fig2}c,f), as is often the case in medical images.    
        To summarize, the coarse explanations provided by ProtoPNet were not informative as already highlighted in \cite{xu2023sanity, gautam2023looks}, while the small receptive field of Proto-BagNet leads to fine-grained and localized explanations.

    \subsection{Proto-BagNet learns meaningful and relevant prototypes} 
        \label{topk_implication}
        \begin{figure}[t]
            \centering
            \includegraphics[width=\textwidth]{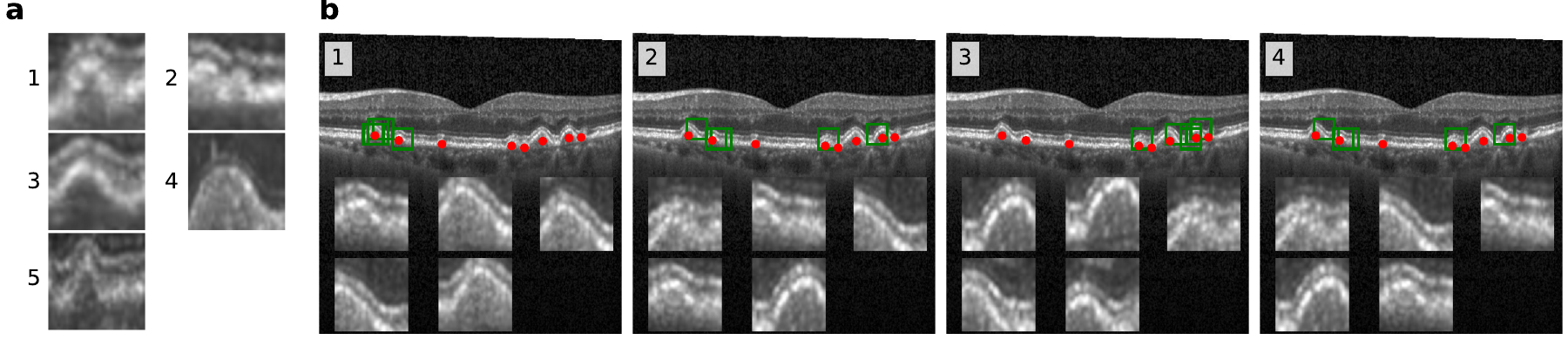}
            %\vspace{-0.9cm}
            \caption{We show \textbf{(a)} the five learned disease prototypes and \textbf{(b)} suspicious regions (green boxes, enlarged below) extracted from each prototype similarity map on an example image. Drusen (annotated with red markers) are detected with high precision.}
            \label{fig3}
            %\vspace{-0.6cm}
        \end{figure}
        
        We assessed the clinical meaning and relevance \cite{ghorbani2019towards, davoodi2023interpretability} of the prototypes learned by Proto-BagNet by evaluating (1) their interpretability and (2) their coherence as the precision of their corresponding similarity maps at localizing drusen. 
        
        We first evaluated the interpretability of the learned prototypes by showing the prototypes without additional context (Fig. \ref{fig3}a) to a clinical expert and asking her if she could understand the concepts encoded in each of them. She described each prototype despite their low resolution as: \textit{(1) soft drusen}; \textit{(2) two drusen in transformation}; \textit{(3) typical drusen}; \textit{(4) drusen with RPE \footnote{Retinal Pigment Epithelium} thinning and dense substance inside}; and \textit{(5) drusen with RPE thinning probably in transformation}, showing that the learned prototype where semantically meaningful even when seen in isolation, and diverse due to the enforced dissimilarity (Suppl. Fig. 1).
        Next, we obtained annotations of the $5$ training images from which the learned prototypes were extracted (Suppl. Fig. 2). 
        We found that all learned prototypes (i.e. $100\%$) were extracted from regions labeled as drusen, confirming that Proto-BagNet learns interpretable and meaningful prototypes. 

        Subsequently, we evaluated the relevance of the learned prototypes on a subset of $40$ test images where an ophthalmologist annotated the presence of drusen. 
        On this subset, we calculated the precision of the prototype similarity maps at localizing drusen lesions to assess whether the highlighted prototypical parts contain similar concepts (drusen related) to those encoded by the learned prototypes. 
        We extracted the $k=5$ prototypical parts as the most discriminative regions similar to each disease prototype (Fig. \ref{fig3}b, more examples in Suppl. Figs. 3,\,4), which were also used in the classification mechanism. 
        From these, we calculated the precision as the proportion of the prototypical parts (green boxes, magnified on the bottom) which contained annotated drusen lesions (red markers) \cite{donteu2023sparse}. 
        The top-$k$ prototypical parts highlighted by the prototype heatmaps contained drusen lesions with high precision ranging from $0.83 \, (k=2)$ to $0.87 \, (k=1)$ depending on $k \in \{1, \ldots, 5\}$, with $p=0.84 \pm 0.2$ (mean $\pm$ SD) for $k=5$.            
    
    \subsection{Proto-BagNet relies on a faithful decision-making process}  
        \label{faithfulness}
\begin{figure}[t]%[b]
            \centering
            \includegraphics[width=\textwidth]{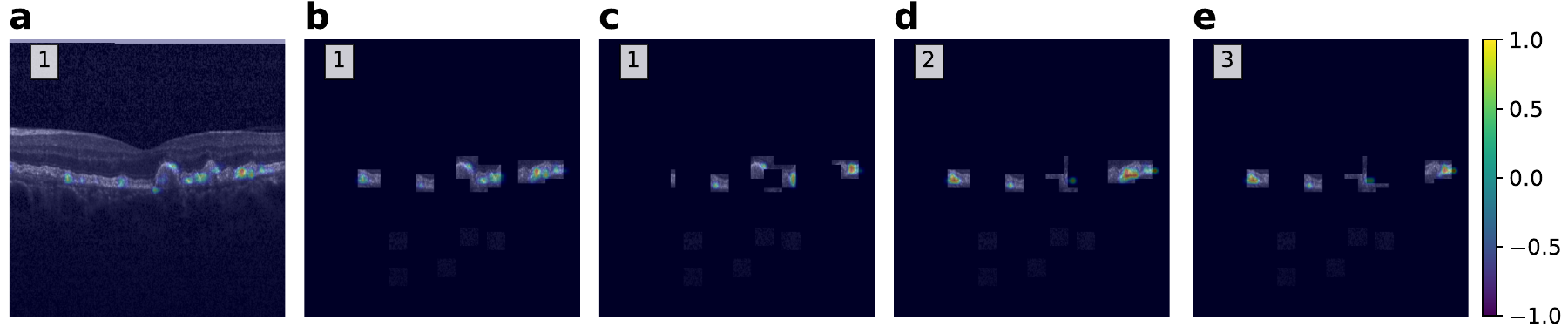}
            %\vspace{-0.9cm}
            \caption{\textbf{(a)} Example drusen image with overlaid prototype similarity map. \textbf{(b)} Occluded image keeping only the top five regions most similar to each prototype. \textbf{(c-e)} Images resulting from occluding the most five regions similar to a prototype (1,2,3).}
            \label{fig4}
            %\vspace{-0.6cm}
        \end{figure}

        Finally, we verified that Proto-BagNet makes decisions based solely on the visual explanation it provides (i.e. on the $k$ input prototypical parts most similar to each prototype). 
        To assess the faithfulness of our model, we applied the model to the test set (example image in Fig. \ref{fig4}a) and then masked all image regions except the top-$k$ prototypical parts identified by the model (Fig. \ref{fig4}b). 
        We reapplied the model to these occluded images and compared the classification output to the output on the original data.
        The distribution of predicted logits on original images ($0.03 \pm 0.06$, and $0.96 \pm 0.15$) was similar to that on occluded images ($0.03 \pm 0.07$, and $0.96 \pm 0.15$), respectively, for healthy and diseased images. 
        The AUC after occlusion was almost similar ($0.9918$ vs $0.9916$) to that obtained without occlusion.
        % In addition, the use of the Mann-Whitney U test did not provide sufficient evidence ($p=0.776$, and $p=0.053$) on healthy and disease images to globally reject the null hypothesis that the predicted distributions are equal with and without occlusion, even though the $p$-value was low for the drusen images.  
        We conclude that Proto-BagNet really makes decisions based only on the prototypical parts, and that the explanations provided by the extracted regions are faithful representations of these prototypical parts.

        We quantified the importance of each prototype by additionally masking its $k$ prototypical parts in the occluded test images (Fig. \ref{fig4}c-e) and measuring the change in classifier predictions.        
        The predicted drusen probability on healthy images increased by $0.27 \pm 0.14$ on average when removing healthy prototypical parts (i.e. the region similar to healthy prototypes), while it decreased by $0.42 \pm 0.17$ when removing disease prototypical parts on drusen images. % see Appendix         
        % Note that occluding an image reduces the regions of interest of the whole image by a factor of $\frac{k \times r^2 \times m \times c}{H_X \times W_X}$ if there are no overlapping patches and even less if not.         
       To summarize, the prototypical parts extracted by the model are indeed evidence for healthy and diseased tissue, respectively, as can be seen by the changes in classifier predictions when occluding them.
        
\section{Discussion and Conclusion}   
    % summary: methods and contribution
    In this work, we proposed Proto-BagNet, an interpretable-by-design prototype-based model that provides faithful and highly localized explanations as well as global interpretability through meaningful prototypes. 
    We evaluated the interpretability of our model through feedback by an ophthalmologist who identified diverse and clinically relevant concepts in the learned prototypes. 
    Furthermore, the model explanations precisely detected drusen lesions in the images. We evaluated the Proto-BagNet for drusen classification on OCT, a solved task in terms of predictive performance (Tab.\,\ref{tab:classification_results}), which lends itself for studying interpretable models due to identifiable and well-characterized regions of interest. 
    However, we noticed that some design choices (e.g., SA layer, k-values) and introduced loss components (dissimilarity and sparsity) enhance interpretability but compete with classification performance (Tab.\,\ref{tab:classification_results}), suggesting that determining the ideal tradeoff might depend on the specific clinical setting. Additionally, the appropriate receptive field size may vary depending on the clinical task and image resolution. In our case, the drusen are small ($< 63 \mu m$ \cite{silvestri2005prevalence}) and fit into a patch of size $33 \times 33$ and can be changed for other tasks to inject clinical knowledge and adjust for resolution.
    In a next step, we believe our approach could also be applied to more challenging tasks, and could be useful for the discovery of unknown relevant concepts. In summary, our work may enable prototype-based networks to take a more central stage for realistic task settings, as a promising alternative to post-hoc explanations of black-box models, in particular on medical images.

\begin{credits}
\subsubsection{\ackname} This project was supported by the Hertie Foundation and the German Science Foundation (Excellence Cluster 2064 “Machine Learning—New Perspectives for Science”, project number 390727645). The authors thank the International Max Planck Research School for Intelligent Systems (IMPRS-IS) for supporting KD.

\subsubsection{\discintname}
The authors declare no competing interests.
\end{credits}

% ---- Bibliography ----
%
% BibTeX users should specify bibliography style 'splncs04'.
% References will then be sorted and formatted in the correct style.
%
\bibliographystyle{splncs04}
\bibliography{Paper-0480}

\begin{thebibliography}{10}
\providecommand{\url}[1]{\texttt{#1}}
\providecommand{\urlprefix}{URL }
\providecommand{\doi}[1]{https://doi.org/#1}

\bibitem{adebayo2018sanity}
Adebayo, J., Gilmer, J., Muelly, M., Goodfellow, I., Hardt, M., Kim, B.: Sanity checks for saliency maps. Advances in neural information processing systems  \textbf{31} (2018)

\bibitem{barnett2021case}
Barnett, A.J., Schwartz, F.R., Tao, C., Chen, C., Ren, Y., Lo, J.Y., Rudin, C.: A case-based interpretable deep learning model for classification of mass lesions in digital mammography. Nature Machine Intelligence  \textbf{3}(12),  1061--1070 (2021)

\bibitem{brendel2019approximating}
Brendel, W., Bethge, M.: Approximating {CNNs} with {Bag}-of-local-{Features} models works surprisingly well on {ImageNet}. In: Proc. International Conference on Learning Representations (ICLR) (2019)

\bibitem{chen2019looks}
Chen, C., Li, O., Tao, D., Barnett, A., Rudin, C., Su, J.K.: This looks like that: deep learning for interpretable image recognition. Advances in neural information processing systems  \textbf{32} (2019)

\bibitem{davoodi2023interpretability}
Davoodi, O., Mohammadizadehsamakosh, S., Komeili, M.: On the interpretability of part-prototype based classifiers: a human centric analysis. Scientific Reports  \textbf{13}(1),  23088 (2023)

\bibitem{fleckenstein2021age}
Fleckenstein, M., Keenan, T.D., Guymer, R.H., Chakravarthy, U., Schmitz-Valckenberg, S., Klaver, C.C., Wong, W.T., Chew, E.Y.: Age-related macular degeneration. Nature reviews Disease primers  \textbf{7}(1), ~31 (2021)

\bibitem{gautam2023looks}
Gautam, S., H{\"o}hne, M.M.C., Hansen, S., Jenssen, R., Kampffmeyer, M.: This looks more like that: Enhancing self-explaining models by prototypical relevance propagation. Pattern Recognition  \textbf{136},  109172 (2023)

\bibitem{ghorbani2019towards}
Ghorbani, A., Wexler, J., Zou, J.Y., Kim, B.: Towards automatic concept-based explanations. Advances in neural information processing systems  \textbf{32} (2019)

\bibitem{He_2016_CVPR}
He, K., Zhang, X., Ren, S., Sun, J.: Deep residual learning for image recognition. In: Proceedings of the IEEE Conference on Computer Vision and Pattern Recognition (CVPR) (June 2016)

\bibitem{huang2023evaluation}
Huang, Q., Xue, M., Huang, W., Zhang, H., Song, J., Jing, Y., Song, M.: Evaluation and improvement of interpretability for self-explainable part-prototype networks. In: Proceedings of the IEEE/CVF International Conference on Computer Vision. pp. 2011--2020 (2023)

\bibitem{jacovi2020towards}
Jacovi, A., Goldberg, Y.: Towards faithfully interpretable {NLP} systems: How should we define and evaluate faithfulness? In: Proceedings of the 58th Annual Meeting of the Association for Computational Linguistics. pp. 4198--4205 (Jul 2020)

\bibitem{kermany2018identifying}
Kermany, D.S., Goldbaum, M., Cai, W., Valentim, C.C., Liang, H., Baxter, S.L., McKeown, A., Yang, G., Wu, X., Yan, F., et~al.: Identifying medical diagnoses and treatable diseases by image-based deep learning. cell  \textbf{172}(5),  1122--1131 (2018)

\bibitem{donteu2023sparse}
Kerol, D., Ilanchezian, I., K{\"u}hlewein, L., Faber, H., Baumgartner, C.F., Bah, B., Berens, P., Koch, L.M.: Sparse activations for interpretable disease grading. In: Medical Imaging with Deep Learning (2023)

\bibitem{kim2021xprotonet}
Kim, E., Kim, S., Seo, M., Yoon, S.: Xprotonet: diagnosis in chest radiography with global and local explanations. In: Proceedings of the IEEE/CVF conference on computer vision and pattern recognition. pp. 15719--15728 (2021)

\bibitem{koh2020concept}
Koh, P.W., Nguyen, T., Tang, Y.S., Mussmann, S., Pierson, E., Kim, B., Liang, P.: Concept bottleneck models. In: International conference on machine learning. pp. 5338--5348. PMLR (2020)

\bibitem{rudin2019stop}
Rudin, C.: Stop explaining black box machine learning models for high stakes decisions and use interpretable models instead. Nature machine intelligence  \textbf{1}(5),  206--215 (2019)

\bibitem{selvaraju2017visual}
Selvaraju, R., Cogswell, M., Das, A., Vedantam, R., Parikh, D., Grad-Cam, B.: Visual explanations from deep networks via gradient-based localization. In: Proceedings of the 2017 IEEE International Conference on Computer Vision (ICCV). pp. 618--626

\bibitem{silvestri2005prevalence}
Silvestri, G., Sillery, E., Henderson, D., Brogan, P., Silvestri, V.: Prevalence of drusen and drusen size in young adults. Investigative Ophthalmology \& Visual Science  \textbf{46}(13),  3298--3298 (2005)

\bibitem{xu2023sanity}
Xu{-}Darme, R., Qu{\'{e}}not, G., Chihani, Z., Rousset, M.: Sanity checks for patch visualisation in prototype-based image classification. In: {IEEE/CVF} Conference on Computer Vision and Pattern Recognition, {CVPR} 2023 - Workshops, Vancouver, BC, Canada, June 17-24, 2023. {IEEE} (2023)

\bibitem{zhou2016learning}
Zhou, B., Khosla, A., Lapedriza, A., Oliva, A., Torralba, A.: Learning deep features for discriminative localization. In: Proceedings of the IEEE conference on computer vision and pattern recognition. pp. 2921--2929 (2016)

\end{thebibliography}

\newpage
\appendix
\setcounter{figure}{0}
\setcounter{table}{0}

\section*{Supplementary material}

        \begin{figure}
            \centering
            \includegraphics[width=\textwidth]{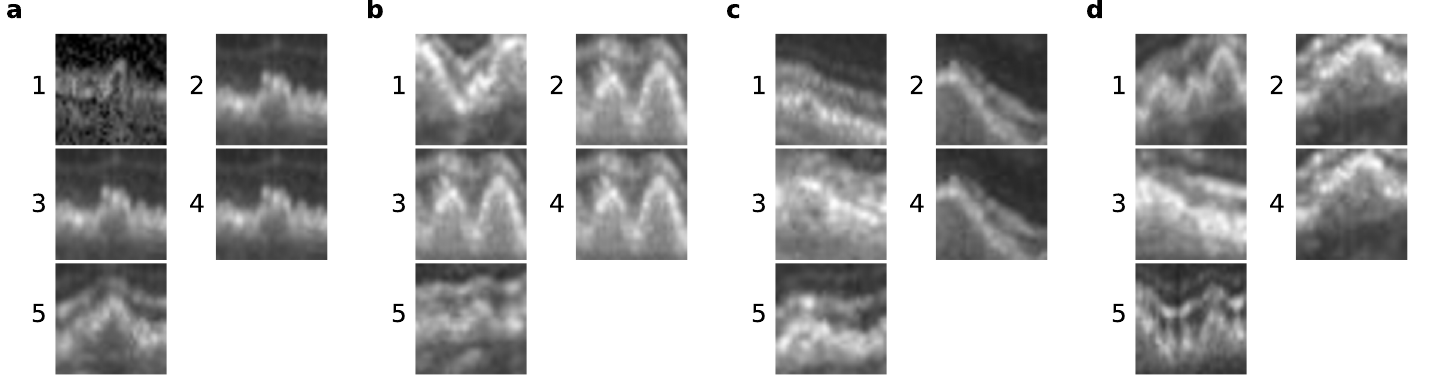}
            %\vspace{-0.5cm}
            \caption{Examples of some learned prototypes without adding the dissimilarity loss to prevent the model from learning redundant prototypes. \textbf{(a,b)} Prototypes 2,3, and 4 are duplicated. \textbf{(c,d)} Prototypes 2 and 4 are duplicated. }
            \label{app_training_img_protoo}
        \end{figure}

        \begin{figure}
            \centering
            \includegraphics[width=\textwidth]{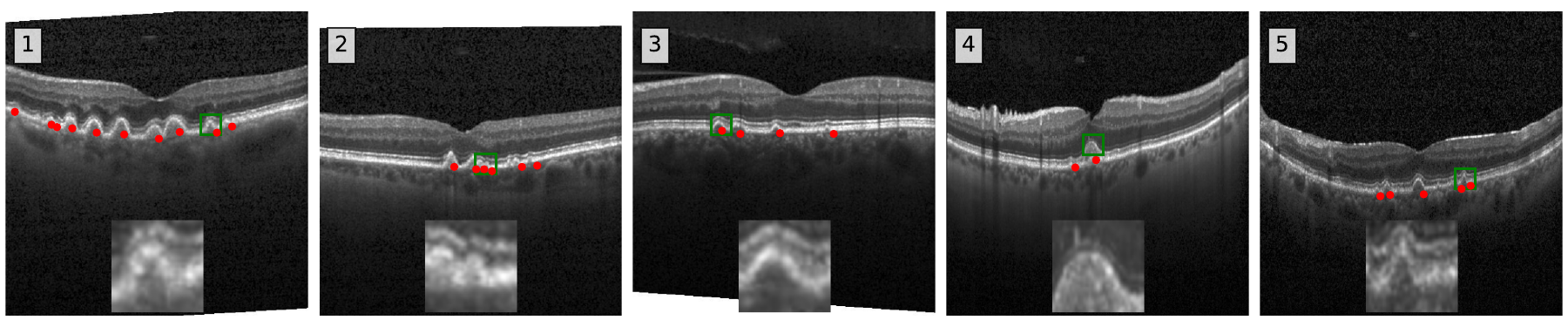}
            \caption{Annotated training images from which the disease prototypes were extracted. The green boxes indicate the region where the learned prototypes were extracted, which are enlarged at the bottom. The red markers denote the reference annotations of drusen lesions. The number at the top indicates the prototype ID. For prototype 4, the bounding box is slightly above the lesion, probably due to a mistake when clicking on the lesion.}
            \label{suppl_fig1}
        \end{figure}

        \begin{figure}
            \centering
            \includegraphics[width=\textwidth]{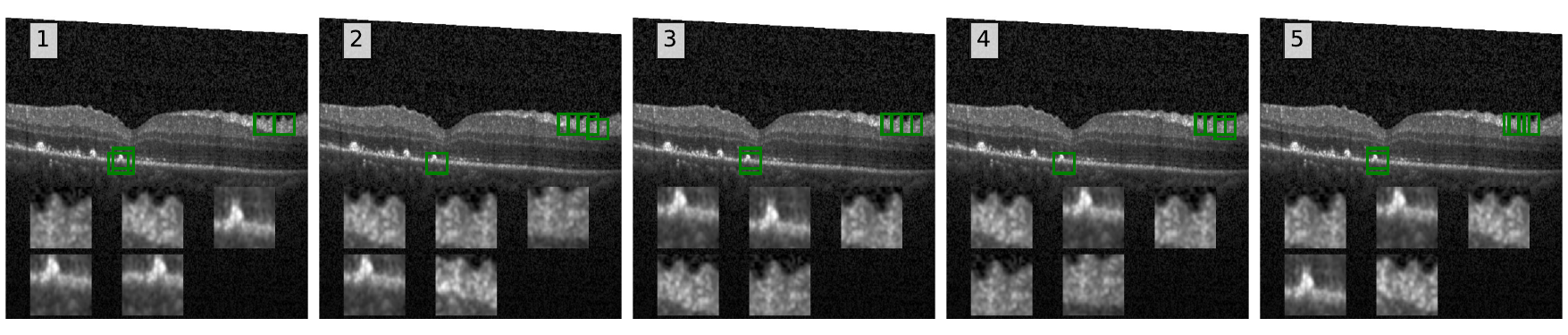}
            \caption{Example of suspicious regions highlighted by Proto-BagNet on a disease image where the ophthalmologist did not find drusen lesions. The region highlighted near the Retinal Pigment Epithelium are sub-retinal deposits which are not typical drusen lesions but wringing of the ganglion cell layer.}
            \label{app_training_img_protoo}
        \end{figure}

    %\section{Additional examples of lesion detection}
        %\vspace{-0.9cm}
        \begin{figure}
            \centering
            \includegraphics[width=\textwidth]{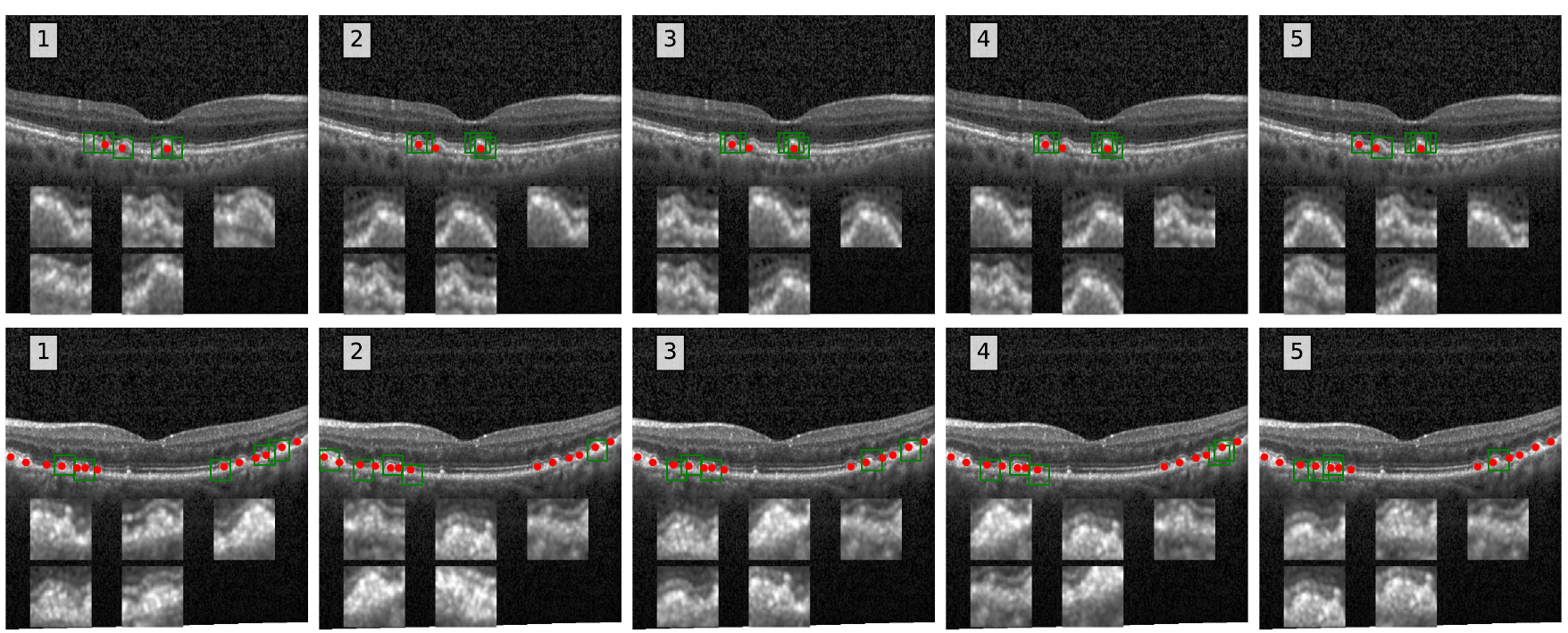}
            \caption{Two examples of suspicious lesions extracted from each prototype similarity map on disease images. Drusen (annotated with red markers) are detected with high precision.}
            \label{app_training_img_proto}
        \end{figure}

        \begin{table}[p]
            \centering
            \caption{Classification performance with confidence intervals (CIs) for drusen detection on validation and test sets. CIs are derived from bootstrapping with n=1000.}
            \begin{tabular}{|l| c| c| c| c| c| c| c| c|}
                \hline
                & \multicolumn{4}{c|}{Validation set} & \multicolumn{4}{c|}{Test set} \\
                 & Accuracy & AUC & Recall & Precision & Accuracy & AUC & Recall & Precision \\
                \hline
                ResNet-50 & $.99 \pm 1e$-$4$ & $.99 \pm 1e$-$4$ & $.98 \pm 2e$-$4$ & $.99 \pm 2e$-$4$ & $.99 \pm 1e$-$4$ & $.99 \pm 1e$-$4$ & $.98 \pm 2e$-$4$ & $.99 \pm 2e$-$4$ \\
                dense BagNet & $.99 \pm 1e$-$4$ & $.99 \pm 1e$-$4$ & $.98 \pm 2e$-$4$ & $.98 \pm 2e$-$4$ & $.99 \pm 1e$-$4$ & $.99 \pm 1e$-$4$ & $.98 \pm 2e$-$4$ & $.98 \pm 2e$-$4$ \\
                \hline
                ProtoPNet & $.99 \pm 1e$-$4$ & $.99 \pm 1e$-$4$ & $.98 \pm 2e$-$4$ & $.97 \pm 2e$-$4$ & $.99 \pm 1e$-$4$ & $.99 \pm 1e$-$4$ & $.98 \pm 1e$-$4$ & $.97 \pm 1e$-$4$ \\
                Proto-BagNet & $.98 \pm 1e$-$4$ & $.99 \pm 1e$-$4$ & $.94 \pm 3e$-$4$ & $.98 \pm 2e$-$4$ & $.98 \pm 1e$-$4$ & $.99 \pm 1e$-$4$ & $094 \pm 3e$-$4$ & $.98 \pm 2e$-$4$ \\
                \hline
            \end{tabular}
            \label{tab:classification_results}
        \end{table}

\end{document}